%% file: main.tex
\documentclass[journal=eds]{CUP-JNL-DTM}%
\usepackage[utf8]{inputenc}

\usepackage{graphicx}
\usepackage{multicol,multirow}
\usepackage{amsmath,amssymb,amsfonts}
\usepackage{mathrsfs}
\usepackage{amsthm}
\usepackage{rotating}
\usepackage{appendix}
\usepackage{ifpdf}
\usepackage[T1]{fontenc}
\usepackage{newtxtext}
\usepackage{newtxmath}
\usepackage{textcomp}
\usepackage{xcolor}
\usepackage{lipsum}
\usepackage[colorlinks,allcolors=blue]{hyperref}
\usepackage{dsfont}

\theoremstyle{definition}

\numberwithin{equation}{section}

\usepackage{float}
\usepackage{enumitem}
\usepackage{bm}
\usepackage{booktabs} 
\usepackage{eurosym}
\usepackage{tabularx} 
\usepackage{adjustbox}
\usepackage{subcaption}
\usepackage{makecell}
\usepackage{tikz}
\usepackage{url}

\usepackage[nameinlink]{cleveref}

\usepackage{framed}
\definecolor{shadecolor}{RGB}{230,230,230}
\definecolor{darkgreen}{RGB}{0,100,0}

\jname{Environmental Data Science}
\articletype{Application Paper}
\jyear{2026}

\graphicspath{{Figures/}}

\renewcommand{\abstractname}{\color{darkgreen}Abstract}

\begin{document}

\begin{Frontmatter}

\title[Article Title]{Cross-Domain Offshore Wind Power Forecasting: Transfer Learning Through Meteorological Clusters}

\author[1]{Dominic Weisser}
\author[2]{Chlo\'e Hashimoto-Cullen}
\author[1,3]{Benjamin Guedj}

\address[1]{\orgname{University College London}, \orgaddress{\country{United Kingdom}}}
\address[2]{\orgname{Sorbonne Université}, \orgaddress{\country{France}}}
\address[3]{\orgname{Inria}, \orgaddress{\country{France}}}


\abstract{\input{Chapters/Abstract}}

\end{Frontmatter}

\begin{shaded}
\noindent{\small\bfseries\color{black}Impact Statement}

\medskip
\noindent This work proposes a novel transfer learning pipeline that achieves accurate offshore wind power forecasts with under five months of local wind farm data. Results demonstrate that early-stage uncertainty and costs can effectively be reduced, avoiding delays to project development.
\end{shaded}

\section{Introduction}
\input{Chapters/Introduction} 

\section{Methodology}
\input{Chapters/Methodology}

\section{Results}
\input{Chapters/Results} 

\section{Discussion}
\input{Chapters/Conclusion}

\begin{Backmatter}
\printbibliography
\end{Backmatter}

\section*{Acknowledgements}
The authors the anonymous reviewers and meta-reviewers for providing constructive and helpful remarks.
Chlo\'{e} Hashimoto-Cullen would like to thank Sylvain Le Corff for proof-reading the manuscript. Chlo\'{e} Hashimoto-Cullen is co-funded by the European Union’s Horizon Europe research and innovation programme Cofund SOUND.AI under the Marie Sklodowska-Curie Grant Agreement No 101081674. Views and opinions expressed are however those of the author(s) only and do not necessarily reflect those of the European Union or the granting authority. Neither the European Union nor the granting authority can be held responsible for them. 

\appendix 
\crefalias{section}{appendix}
\section{Empirical Details}
\label{app:empirical_details}
\input{Chapters/AppendixA}

\section{Metrics Used}
\label{app:metric_def}
\input{Chapters/AppendixB}

\section{Optimal Configuration Selection} \label{app:config_selection}
\input{Chapters/AppendixC}

\section{Baseline Results} 
\label{app:baseline_results}
\input{Chapters/AppendixD}

\end{document}

%% file: Chapters/Abstract.tex
Ambitious decarbonisation targets are rapidly increasing the commission of new offshore wind farms. For these newly commissioned plants to run, accurate power forecasts are needed from the onset. These allow grid stability, good reserve management and efficient energy trading. Despite machine learning models having strong performances, they tend to require large volumes of site-specific data that new farms do not yet have. To overcome this data scarcity, we propose a novel transfer learning framework that clusters power output according to covariate meteorological features. Rather than training a single, general-purpose model, we thus forecast with an ensemble of expert models, each trained on a cluster. As these pre-trained models each specialise in a distinct weather pattern, they adapt efficiently to new sites and capture transferable, climate-dependent dynamics. Our contributions are two-fold - we propose this novel framework and comprehensively evaluate it on eight offshore wind farms, achieving accurate cross-domain forecasting with under five months of site-specific data. Our experiments achieve a MAE of 3.52\%, providing empirical verification that reliable forecasts do not require a full annual cycle. Beyond power forecasting, this climate-aware transfer learning method opens new opportunities for offshore wind applications such as early-stage wind resource assessment, where reducing data requirements can significantly accelerate project development whilst effectively mitigating its inherent risks.

%% file: Chapters/Introduction.tex
Offshore wind power is tapped to be one of the cornerstones of the green energy transition, with global offshore wind capacity projected to triple from 83~GW in 2024 to 238~GW by 2030 \citep{ember2025wind}. However, current wind farms are concentrated in a small number of developed countries \citep{statista_offshore_wind_farms}. This leaves many emerging market countries unexploited, with an estimated potential of 3.1~TW of offshore wind power between Brazil, India, Morocco, the Philippines, South Africa and Vietnam \citep{WorldBank_OffshoreWind_Emerging_Markets_2019}. Of these, Vietnam is currently the only one with offshore wind farms \citep{GWEC_Global_Wind_Report_2025}.

To exploit these markets, both efficient Wind Resource Assessment (WRA) and accurate Wind Power Forecasting (WPF) from the start of operations are needed. In nascent markets with the highest uncertainty, reliable WPF is crucial to integrating intermittent generation into local grids, as it provides system stability, limits reserve requirements and hedges financial risks.

Traditional physical models for WPF are robust but are computationally intensive and highly sensitive to initial parametrisation \citep{en18020350}. Consequently, statistical and machine learning approaches have gained in prominence; by learning the complex, non-linear relationships between meteorological variables and power output directly from historical data, these data-driven models effectively circumvent the computational overhead required by purely physical simulations. Among these, deep learning (DL) approaches have state-of-the-art (SOTA) performance in offshore WPF \citep{Ally2025Modular}, in particular due to their abilities to model high-dimensional dependencies. However, this success relies heavily on granular, site-specific SCADA data. Consequently, the data-intensive nature of current SOTA frameworks inherently restricts their real-world deployment for newly commissioned sites and emerging markets where historical records are scarce \citep{JIN2025138275}.

Transfer learning is therefore emerging as an approach to leverage existing knowledge from a data-rich source domain $\mathcal{D}_S$ to a data-scarce target domain $\mathcal{D}_T$ \citep{5288526}. Current approaches transfer farm-level models \citep{10582569, LiDanHu}, without exploiting possible structure in the meteorological inputs. However, identifying this structure and training individual models for each identified weather pattern can improve WPF performance \citep{7727853}. To the best of our knowledge, these approaches have not yet been used to enable cross-site knowledge transfer.

\begin{figure}[ht]
    \centering
    \includegraphics[width=0.85\textwidth]{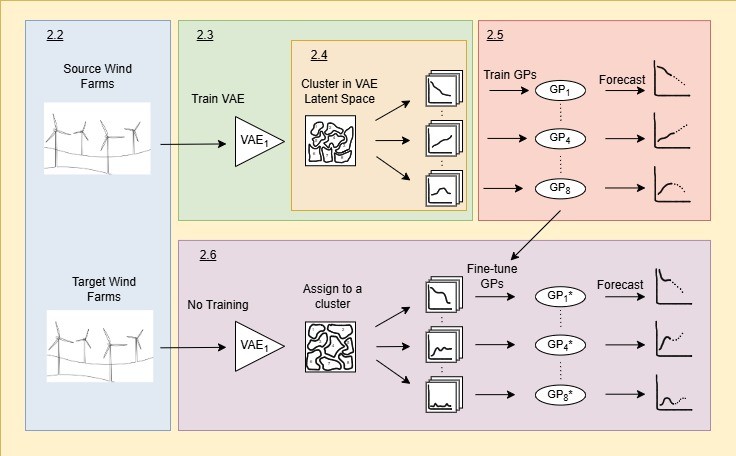}
    \caption{Methodology overview.}
    \label{fig: Overview of our methodology.}
\end{figure}

In this paper, we propose a transfer learning framework that bridges this gap. By investigating whether transfer learning with limited historical data across various European sites can achieve high performance, we thus emulate the constraints of newly commissioned sites. Our model learns domain-agnostic representations of offshore weather across Northern Europe, which can be clustered by meteorological similarity rather than location. Thus, one-hour ahead forecasting models can be trained for each weather condition, and data from target wind farms can be mapped by similarity to an existing meteorological cluster. The relevant model can then be fine-tuned with minimal site-specific data. We present our global methodology in \Cref{fig: Overview of our methodology.}.

The remainder of this paper is organised as follows. \Cref{sec:methodo} describes the dataset and methodology used. \Cref{sec:results} presents and analyses the experimental results, for which \Cref{sec:discussion} provides further discussion\footnote{The full code is available at \url{https://github.com/DomWeisser/offshore-wind-forecasting-2026}}.

%% file: Chapters/Methodology.tex
\label{sec:methodo}

\subsection{Dataset}
We use the dataset introduced by \citet{grothe2022analyzing}, comprising 40 years of hourly meteorological and synthetic power data for 29 European offshore wind farms. The meteorological data is derived from ERA5, a global reanalysis dataset produced by the European Centre for Medium-Range Weather Forecasts (ECMWF) that provides consistent hourly estimates of certain atmospheric variables. By combining the ERA5 reanalysis data with turbine-specific power curves, the authors generate realistic hourly power production time series, designed for research purposes. While this dataset relies on generated power curves and grid-averaged reanalysis data rather than high-resolution operational SCADA measurements, it provides a controlled and consistent baseline across all target farms. Avoiding the missing values, sensor errors and turbine degradation typical of real-world SCADA data allows us to isolate the theoretical efficacy of our proposed transfer learning framework.

\subsection{Selecting Source Wind Farms}
To capture a comprehensive representation of meteorological patterns across Northern Europe, source wind farms are selected to maximise meteorological and operational diversity. Exposure to such diversity allows the models to learn from a broader range of atmospheric patterns, enhancing their ability to generalise to new sites that have distinct meteorological characteristics.

For each of the 29 available wind farms, a 38 dimensional vector is extracted from data recorded between January 1, 2018, and December 31, 2019. The feature set included wind speed statistics (mean, variance, percentiles), sea surface roughness distributions, wind directional consistency, power output metrics (including capacity factor and efficiency) and the frequency of extreme weather events. Dimensionality reduction is performed using Uniform Manifold Approximation and Projection \citep{mcinnes2020umapuniformmanifoldapproximation}, configured to consider 15 nearest neighbours for each data point and with a minimum distance of 0.1. This is followed by agglomerative hierarchical clustering \citep{Ward1963}, which partitions the farms into six distinct clusters based on both their meteorological and operational profiles. From each cluster, one farm is selected to represent the cluster as a source wind farm.

\begin{figure}[ht]
    \centering
    \begin{subfigure}[b]{0.49\linewidth}
        \centering
        \includegraphics[width=\linewidth, trim={1.5cm 1.5cm 1.5cm 1.5cm}, clip]{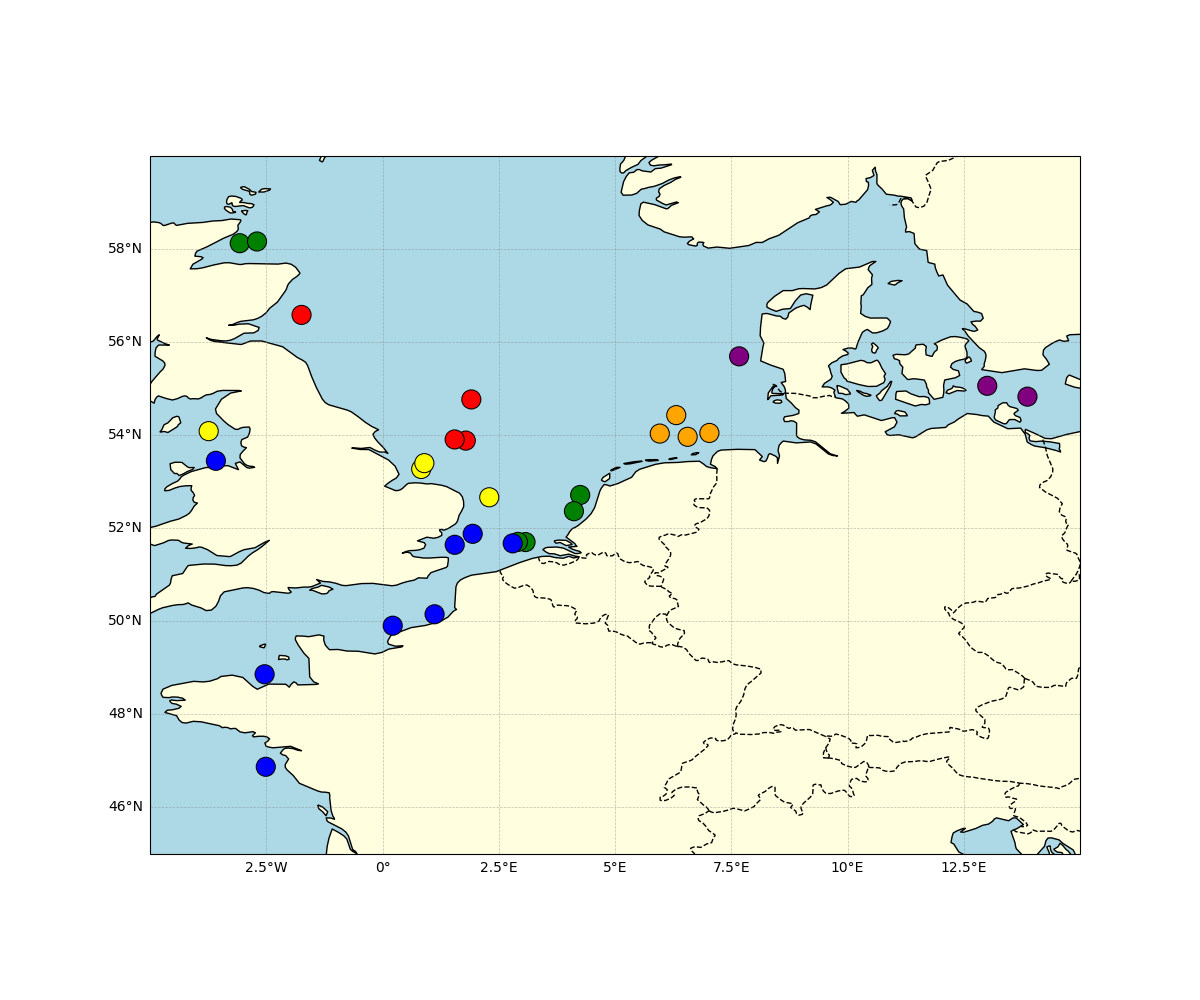}
        \caption{Distribution of wind farm clusters.}
        \label{wind_farm_clustering_umap.png}
    \end{subfigure}
    \hfill
    \begin{subfigure}[b]{0,49\linewidth}
        \centering
        \includegraphics[width=\linewidth, trim={1.5cm 1.5cm 1.5cm 1.5cm}, clip]{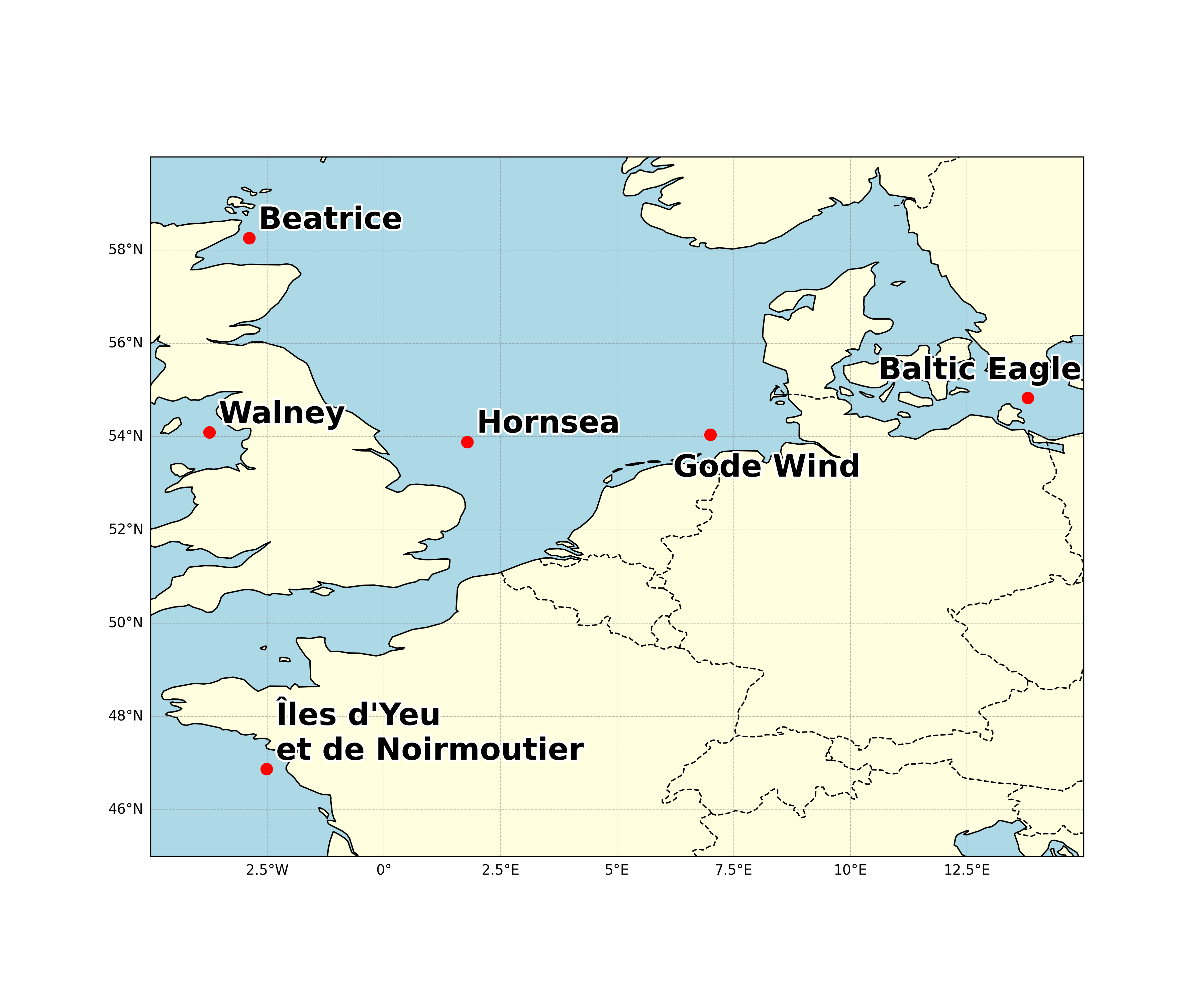}
        \caption{Wind farms selected for source models.}
        \label{fig:Target Wind Farms}
    \end{subfigure}
    \caption{Location and cluster assignment for available wind farms with the selected source wind farms.}
\end{figure}

\subsection{Identifying and Clustering Weather Patterns}

To identify distinct weather patterns across the six source wind farms, hourly meteorological data from all farms is segmented into continuous, non-overlapping time periods of length $p$ hours. Each period $i$ is represented as a multivariate time-series $\mathbf{X}_i \in \mathbb{R}^{p \times d}$, with $d=6$ meteorological features: wind speed, sea surface roughness, wind direction components ($\sin\theta$ and $\cos\theta$ for a wind direction $\theta$) and horizontal and vertical wind velocity components at 100m above sea level ($u_{100}$, $v_{100}$). To ensure the forecasting models are trained on the most distinct weather patterns, the time-period length $p$ and cluster count $K$ are optimised via grid search, maximising the separation between identified weather patterns (see \Cref{sec:2.4}).

For a fixed time-period length $p$, we train a Variational Autoencoder (VAE) \citep{kingma2014auto} to learn a compressed representation of each weather pattern, which we model as a Gaussian distribution $q_\phi(\cdot\mid\mathbf{X})$ with mean $\mu_\phi(\mathbf{X})$ and variance $\sigma_\phi(\mathbf{X})$. It simultaneously learns a decoder, a Gaussian distribution $p_\theta(\cdot \mid \mathbf{z})$ with mean $\mu_\theta(\mathbf{z}) = \mathbf{\hat{X}}$ and fixed variance to perform the inverse reconstruction.
The encoder $q_\phi$ is parametrised with an architecture containing two 1D convolutional layers with kernel sizes of 5 and 3, designed to extract short-term local features. These feature maps are processed by a Bidirectional Long Short-Term Memory (BiLSTM) network, which captures global temporal dependencies by traversing the sequence in both forward and backward directions \citep{650093}. The final concatenated hidden states are projected via fully connected layers to parametrise the variational posterior, $\mu_\phi(\mathbf{X}) \in \mathbb{R}^{8}$ and $\log \sigma_\phi^{2}(\mathbf{X}) \in \mathbb{R}^{8}$. The input multivariate time series are thus mapped to latent variables that are an 8-dimensional latent vector $\mathbf{z} \in \mathbb{R}^{8}$. The latent variables are subsequently sampled via the reparametrisation trick $\mathbf{z} = \mu_\phi(\mathbf{X}) + \sigma_\phi(\mathbf{X}) \odot \boldsymbol{\varepsilon}$, where $\boldsymbol{\varepsilon} \sim \mathcal{N}(\mathbf{0}, \mathbf{I})$ and $\odot$ denotes element-wise multiplication.
The decoder, $p_\theta(\mathbf{X} \mid \mathbf{z})$, performs the inverse reconstruction by employing a unidirectional LSTM for sequence generation.
The latent vector $\mathbf{z}$ is first expanded via a fully connected layer and repeated across the temporal dimension $p$ to initialise the sequence. This sequence is processed by the unidirectional LSTM to reconstruct the temporal dynamics, followed by transposed convolutional layers to recover the original multivariate dimensions and local structure of the time-series $\hat{\mathbf{X}}$. The architecture of the model is inspired by \citet{mehmood2026accurate}.

The model parameters $\theta$ and $\phi$ are optimised by minimising the $\beta$-VAE loss function \citep{higgins2017betavae} with mean squared error reconstruction loss and KL divergence regularisation:
\begin{equation*}
\mathcal{L}(\theta, \phi; \mathbf{X}) =
\mathbb{E}_{q_\phi(\mathbf{z} \mid \mathbf{X})}
\Big(\lVert \mathbf{X} - \hat{\mathbf{X}} \rVert_2^2\Big)
+ \beta \,
D_{\mathrm{KL}}
\Big(
q_\phi(\mathbf{z} \mid \mathbf{X})
\;\|\;
\mathcal{N}(\mathbf{0}, \mathbf{I})
\Big).
\end{equation*}
Training was run for up to 200 epochs using AdamW optimisation \citep{loshchilov2019decoupledweightdecayregularization}. To mitigate posterior collapse, we employ a monotonic annealing schedule for $\beta$, linearly increasing from $0.01$ to $1.0$. For inference, each period was deterministically encoded as the variational posterior mean, $\mathbf{z}_i = \mu_\phi(\mathbf{X}_i)$ providing a concise representation of the meteorological temporal dynamics. The latent vectors are L2-normalised to prioritise directional similarity over magnitude, then partitioned into $K$ distinct weather clusters using Hierarchical Agglomerative clustering with Ward's linkage \citep{Ward1963} and Euclidean distance.

\subsection{Optimal Configuration Selection}
\label{sec:2.4}
The optimal configuration ($p$, $K$) is selected via grid search by maximising a composite quality score $Q$. Drawing upon the multi-objective framework of \textcite{7727853}, the composite score aggregates six metrics, designed to balance statistical rigour in the latent space with physical interpretability and temporal coherence:
\begin{equation*}
    Q = 0.2 S + 0.2 \tilde{D} + 0.2 \tilde{C} + 0.2 \tilde{M} + 0.1 T + 0.1 H.
\end{equation*}
Structural quality is evaluated using the Silhouette score $S$ \citep{rousseeuw1987silhouettes}, which measures cluster cohesion and separation; the normalised Davies-Bouldin index $\tilde{D}$ \citep{Davies1979ACS} which evaluates intra-cluster similarity relative to inter-cluster separation; and the normalised Calinski-Harabasz index $\tilde{C}$ \citep{Calinksi} which quantifies the ratio of between cluster to within-cluster dispersion. Physical interpretability is validated using meteorological separability $\tilde{M}$, calculated as the log-normalised mean ANOVA F-statistic across wind speed, wind variability and power output while robustness is assessed through temporal coherence $T$, defined as the average within-cluster autocorrelation \citep{WARRENLIAO20051857}, and distributional consistency ($H = 1 - \textrm{JSD}$), which uses the Jennen-Shannon Divergence \citep{Lin1991} to penalise shifts in cluster proportions between training and testing sets. As detailed in \Cref{tab:all_metrics}, a sensitivity analysis reveals a trade-off between these objectives: while structural metrics ($S$, $\tilde{D}$, $\tilde{C}$) prefer compact representations with fewer clusters, physical and temporal metrics ($\tilde{M}$, $T$) improve with more clusters. To prioritise identifying distinct weather clusters, the composite weights were heuristically calibrated into two tiers. Primary structural and meteorological metrics ($S$, $\tilde{D}$, $\tilde{C}$, $\tilde{M}$) drive the core clustering behaviour and were assigned dominant weights (0.2 each). Meanwhile, temporal coherence and distributional consistency ($T$, $H$) act a secondary regularisation terms to prevent fragmented or unstable clusters, receiving lower weights (0.1 each). These generally pointed to similar results when clustering without a composite score, thus why they were combined to agree on final cluster parameters.

\subsection{Gaussian Process Models}

Following the assignment of meteorological periods to the $K$ identified weather clusters, we model the relationship between atmospheric conditions and power generation using cluster-specific Gaussian Process regression \citep{williams2006gaussian}. GPs are particularly advantageous for tackling our "cold-start" problem because they operate non-parametrically and are highly data efficient, allowing them to adapt dynamically to limited target data. Crucially, unlike standard deterministic models, GPs inherently generate well-calibrated predictive distributions. For offshore WPF, this uncertainty quantification is an operational necessity; providing grid operators with the reliable risk margins required to securely manage reserve requirements and maintain system stability \citep{pinson2013wind}. For each weather cluster $k$, a separate GP $f_k(\mathbf{x}) \sim \mathcal{GP}(m_k(\mathbf{x}), k_k(\mathbf{x}, \mathbf{x}'))$ is trained to learn the cluster-dependent power curve, enabling each model to specialise on the distinct dynamics characterising that specific weather type.

To capture both the overall period trajectory and immediate local conditions, we construct a 20-dimensional input vector $\mathbf{x} \in \mathbb{R}^{20}$ for each hourly forecast. This vector concatenates the 8-dimensional latent representation $\mathbf{z}$ with 12 physical predictors: current wind speed and sea surface roughness; autoregressive lags $(t-1, t-2)$ for both power and wind speed; and cyclical encodings ($\sin, \cos$) of the hour, month and wind direction.

To map this input to the target power output, the covariance function (kernel) must be tailored to the multi-scale physical reality of wind power generation. Standard applications of GPs in WPF frequently utilise the infinitely differentiable Squared Exponential (RBF) kernel, which effectively captures the smooth, theoretical aerodynamic relationship between wind speed and power \citep{ROGERS20201124}. However, real-world wind power curves contain high-frequency stochastic variations induced by local turbulence and wake effects, for which recent literature highlights the one-differentiable Matérn-$\nu=3/2$ is better suited \citep{Prakash02012023}. To explicitly capture both phenomena, we construct a additive composite kernel. The RBF component isolates the smooth, macroscopic baseline of the ideal power curve, while the Matérn-$\nu=3/2$ component captures the rougher, unpredictable atmospheric fluctuations:

\begin{equation*}
k_k(\mathbf{x}, \mathbf{x}') = \sigma_{f, k}^2 \left(
    k_{\text{RBF}}(\mathbf{x}, \mathbf{x}'; \boldsymbol{\ell}_k^{\text{RBF}}) +
    k_{\text{Matérn}}(\mathbf{x}, \mathbf{x}'; \boldsymbol{\ell}_k^{\text{Mat}}, \nu=1.5)
\right),
\end{equation*}
where $\sigma_{f,k}^2$ controls the output variance. Both components utilise Automatic Relevance Determination (ARD), where $\boldsymbol{\ell}_k \in \mathbb{R}^{20}$ represents a vector of length-scales. This allows the GP to automatically determine the relevance of each input feature for each specific cluster, for example weighting lags more heavily in stable conditions versus turbulence features during storms. 

Observation noise is modelled via a Gaussian likelihood with variance $\sigma_{n, k}^2$, assuming $y=f_k(\mathbf{x}) + \varepsilon$ where $\varepsilon \sim \mathcal{N}(0, \sigma_{n, k}^2)$. Consequently, the covariance of the noisy observations is given by $\mathbf{K}_{y,k} = k_k(\mathbf{x}, \mathbf{x}') + \sigma_{n, k}^2 I$. The complete hyperparameter set $\theta_k = \{\sigma_{f,k}^2, \sigma_{n,k}^2, \boldsymbol{\ell}_k^{\text{RBF}}, \boldsymbol{\ell}_k^{\text{Mat}}, c_k\}$ is optimised by maximising the exact Marginal Log-Likelihood (MLL) of the observations. Using the noisy covariance $\mathbf{K}_{y,k}$, the MLL is defined as:
\begin{equation*}
    \log p(\mathbf{y} \mid X, \theta_k) = -\frac{1}{2}\big( \mathbf{y}^\top \mathbf{K}_{y,k}^{-1} \mathbf{y}\big) - \frac{1}{2} \log |\mathbf{K}_{y,k}| - \frac{N}{2} \log(2\pi).
\end{equation*}

\subsection{\textit{Transfer Learning}}
To evaluate the cross-site generalisation capability of the proposed framework, we apply a transfer learning strategy where the library of GP models are adapted to eight unseen target farms. This adaptation utilises the fixed VAE encoder $\phi_S$ and the set of cluster-specific GP priors, proceeding in three stages: latent projection, cluster alignment and GP fine-tuning. First, the target meteorological time series $\mathbf{X}_t$ are projected into the source latent space using the fixed source encoder $\mathbf{z}_T = {\mu}_{\phi_S}(\mathbf{X}_T)$. This enforces that target weather patterns are interpreted through the learned representation of the source domain. Each target period is then assigned to one of the $K$ source weather clusters via a nearest-neighbourhood classification in the latent space, utilising the centroid from the source training set. 

For each identified weather pattern $c \in \{1, \dots, k\}$, we initialise a target GP $f_T^{(c)}$ using the hyperparameters of the corresponding source GP, $f_S^{(c)}$. This initialisation provides a strong prior on the temporal covariance structure of the power output. The kernel hyperparameters ${\theta}_k$ are then optimised by maximising the exact marginal log-likelihood on the available target training data $\mathcal{D}_{\text{train}}$:
\begin{equation*}
    \hat{{\theta}}_T = \underset{{\theta}}{\mathrm{argmax}} \log p(\mathbf{y}_{\text{train}} \mid \mathbf{X}_{\text{train}}, {\theta}, \sigma^2_{n, \text{fixed}}).
\end{equation*}
To assess data efficiency, we conducted experiments varying the data availability of the two-year target datasets. Each target dataset is split into training fractions $\gamma \in \{0.1, 0.2, 0.3, 0.4, 0.5\}$, simulating a `cold start' scenario where a newly commissioned wind farm has been operational for only a short period (ranging from approximately 2.4 to 12 months). The remaining $1-\gamma$ is reserved for testing.

%% file: Chapters/Results.tex
\label{sec:results}
To evaluate the performance of our transfer learning framework under each data availability scenario, we benchmark our results against three no-transfer baselines. These are individual target farm-level GP and Neural Network (NN) models (experimental details are given in \Cref{app:empirical_details}), which enable comparison with industry standards; and cluster-specific GPs trained from scratch, which helps to quantify the gain from fine-tuning the pre-trained models. We assess the predictive performance of these models using a combination of deterministic and probabilistic metrics \citep{xie2023overview}. Point predictions are evaluated using Mean Absolute Error (MAE) and the coefficient of determination (\( R^2 \)), uncertainty is quantified with the Continuous Ranked Probability Score (CRPS) and 95\% Prediction-Interval Coverage Probability (PICP$_{95}$). The formulas for the metrics are given in \Cref{app:metric_def}. To ensure comparability across wind farms, MAE and CRPS are expressed as a percentage of total capacity, while overall aggregated results are weighted by the number of time-periods in each cluster to account for varying cluster sizes.

\begin{figure}[ht]
    \centering
    \begin{minipage}[b]{0.43\linewidth}
        \centering
        \includegraphics[width=\linewidth]{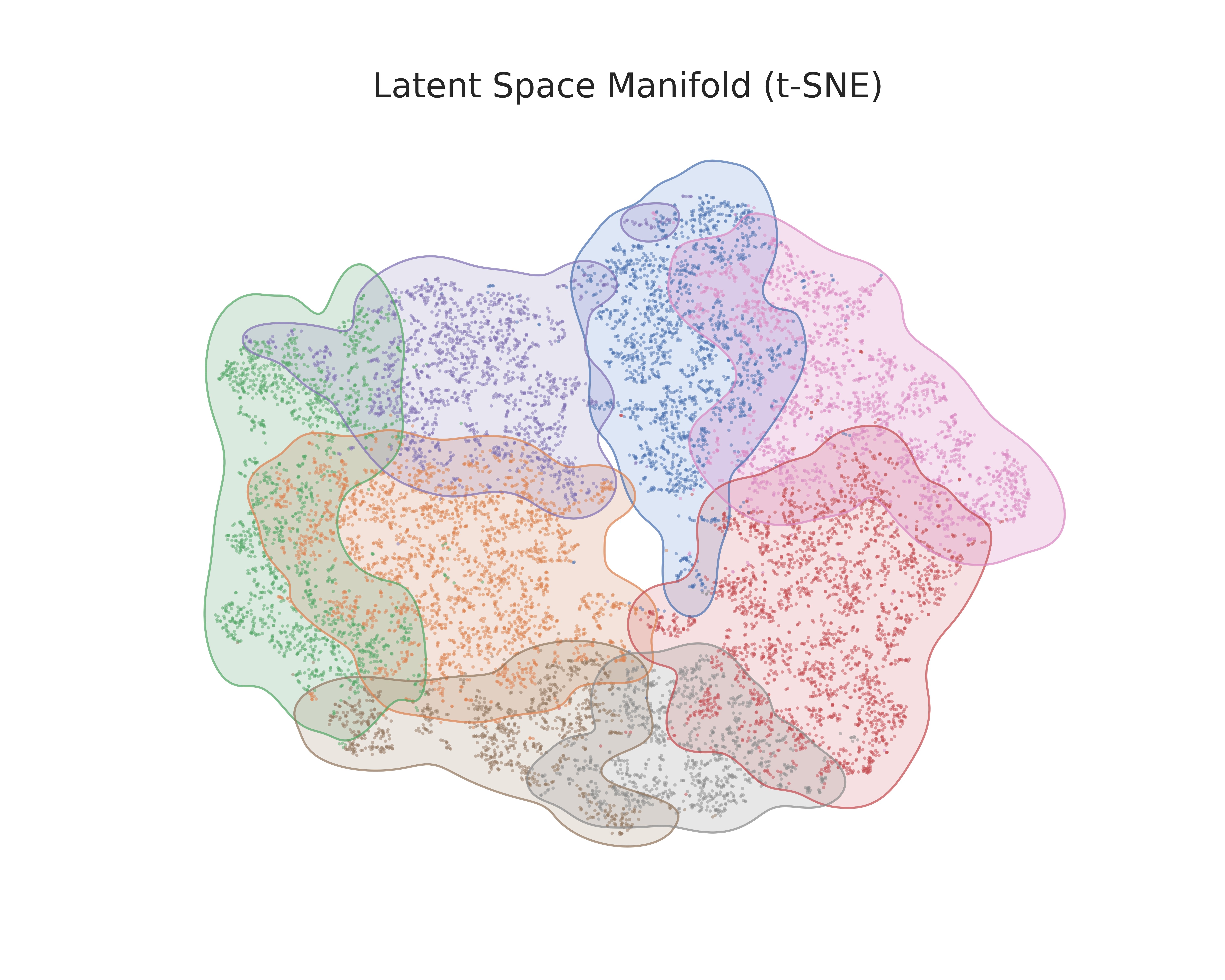}
        \caption{Latent space manifold visualised via t-SNE, showing the distribution of training periods across weather clusters.}
        \label{fig:Periods_split_by_Cluster}
    \end{minipage}
    \hfill
    \begin{minipage}[b]{0.54\linewidth}
        \centering
        \includegraphics[width=\linewidth]{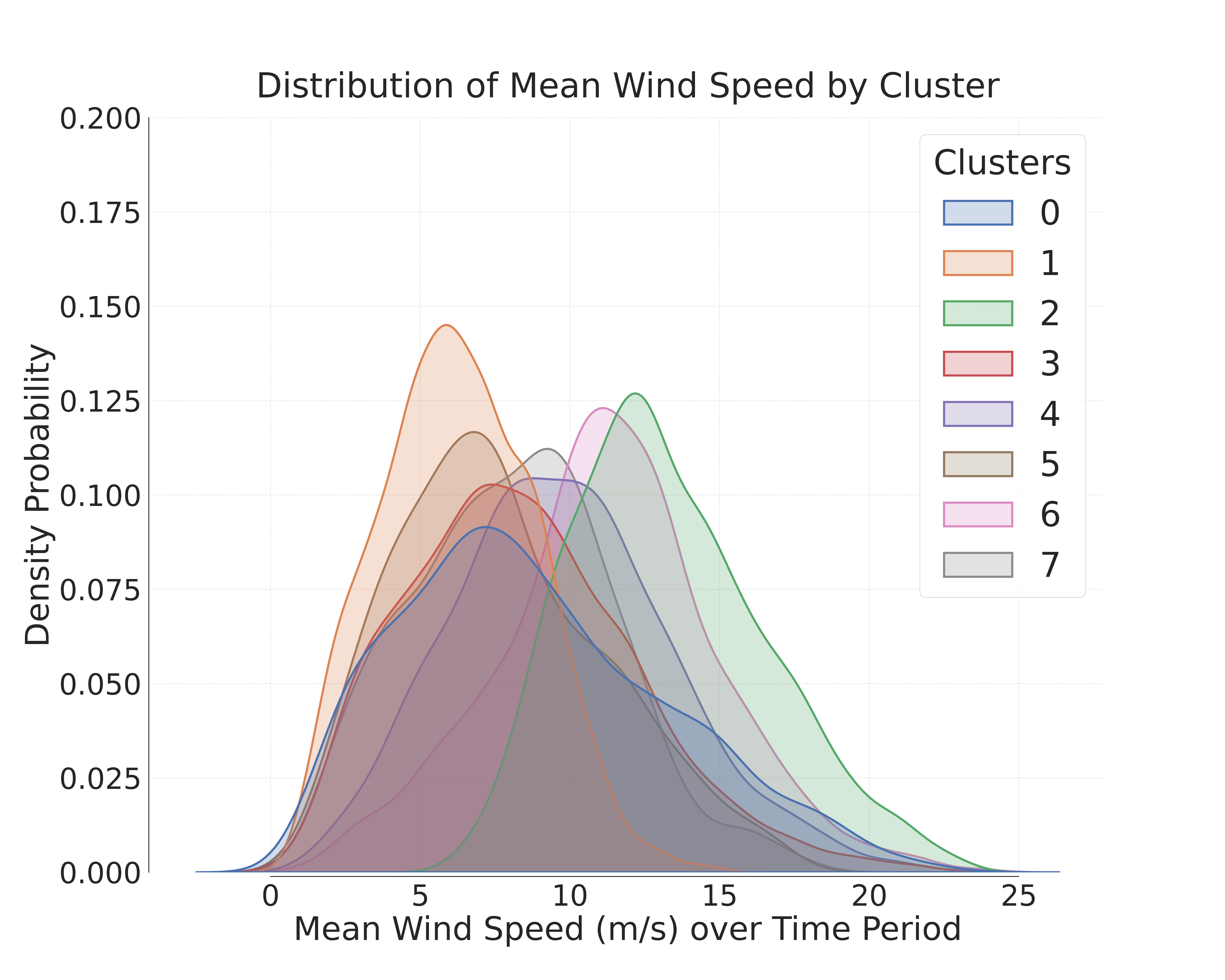}
        \caption{Mean wind speed distribution of training periods by cluster (exemplified by seed 84).}
        \label{fig:ClusterAnalysis}
    \end{minipage}
\end{figure}

The time window with $p=6$ hours and $K=8$ clusters was identified as the best configuration, with quality score of $Q=0.53$. This provides the best trade-off, with cohesive clusters each with distinct meteorological patterns. \Cref{fig:Periods_split_by_Cluster} shows the structural separation, where the eight-dimensional latent representation per training period are embedded into a two dimensional plane using t-SNE \citep{JMLR:v9:vandermaaten08a}. \Cref{fig:ClusterAnalysis} illustrates the physical distinctiveness via mean wind speed distributions of each cluster: Clusters 1 and 5 capture calm periods, Cluster 2 isolates high-speed extremes and Cluster 0 exhibits higher internal variance (likely characterising transitional atmospheric states).

\noindent\textbf{Quantitative Results}
The GP models achieved an MAE of $3.68\% \pm 0.04\%$ (24.4MW; $R^2=0.981$) when evaluated using two years of operational data across the six source farms (4:1 split). 
\Cref{tab:performance} shows the variability of performance for each cluster. \Cref{fig:forecast} shows that this is due to the specificity of each wind farm. Thus, the mixture of GPs captures a rich distribution of possible behaviours, that can be flexible when adapted to new wind farms.
\begin{table}[ht]
\centering
\begin{minipage}{0.40\linewidth}
\centering
\small
\begin{tabular}{lcc}
\toprule
\textbf{Cluster} & \textbf{Test samples} & \textbf{MAE} \\
\midrule
0 & 2682 & 3.93 (0.19) \\
1 & 2845 & 3.38 (0.35) \\
2 & 2110 & 3.32 (0.46) \\
3 & 2228 & 3.76 (0.09) \\
4 & 2187 & 3.97 (0.22) \\
5 & 2170 & 3.44 (0.59) \\
6 & 1290 & 3.90 (0.22) \\
7 & 1312 & 3.78 (0.26) \\
\midrule
\textbf{Overall} & \textbf{17,490} & \textbf{3.68 (0.04)}  \\
\bottomrule
\end{tabular}
\vspace{0.5ex}
\captionof{table}{Cluster-level performance of source GP models, mean (std) across all target farms}
\label{tab:performance}
\end{minipage}
\hfill
\begin{minipage}{0.53\linewidth}
\centering
\includegraphics[width=\linewidth]{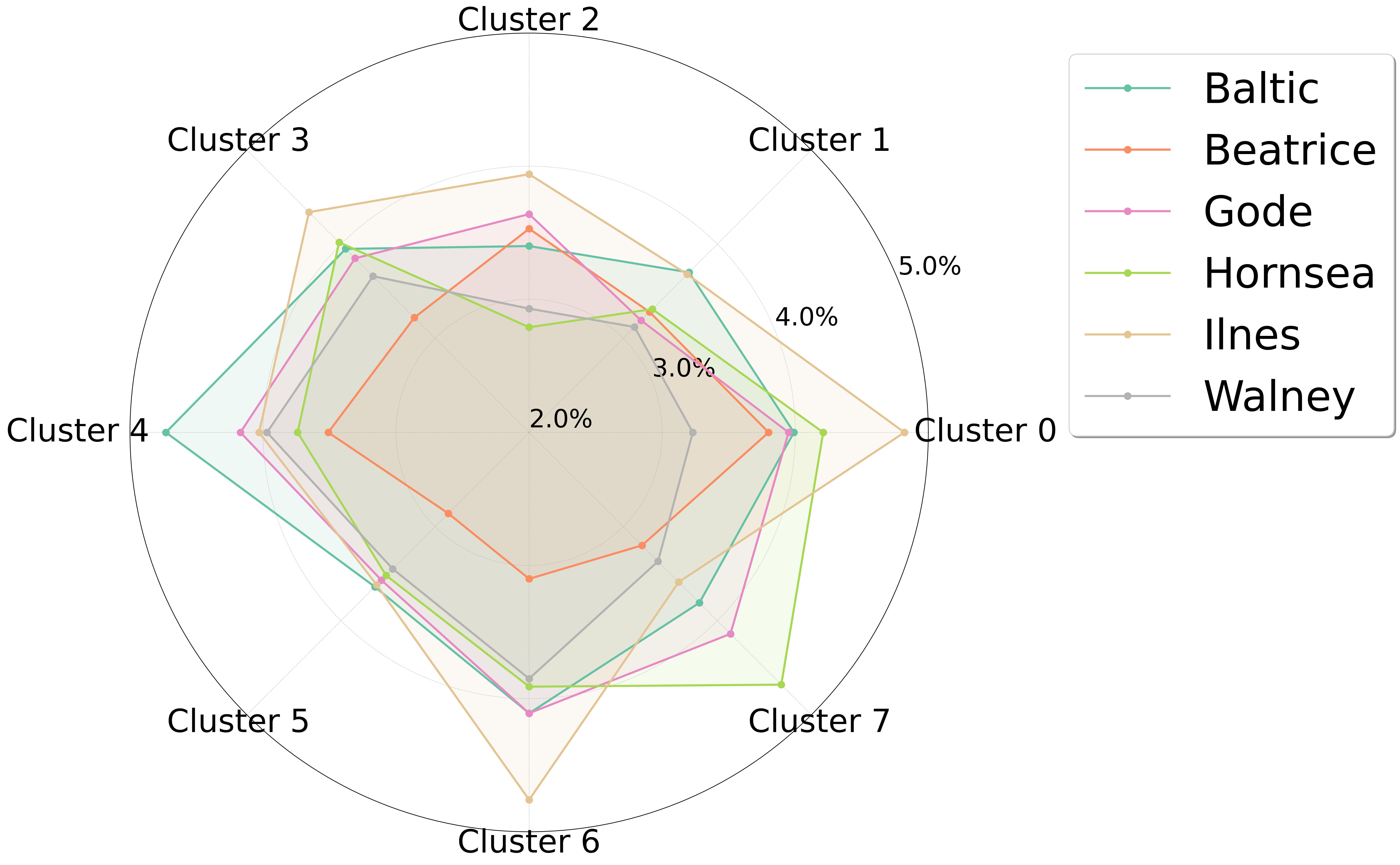}
\vspace{0.5ex}
\captionof{figure}{Performance of GP forecasts, disaggregated by wind farm and cluster.}
\label{fig:forecast}
\end{minipage}
\end{table}

A deterministic evaluation of the transfer learning approach for each cluster and proportion of available data is presented in \Cref{tab:capacity_results_aggregated}, alongside baseline results. The metrics are averaged across the eight target farms. Fully disaggregated baseline results and probabilistic metrics are detailed in \Cref{app:baseline_results}. As expected, when fine-tuning the pre-trained GP models we achieve superior generalisation compared to training cluster-specific GP models from scratch. Notably, our framework requires under five months of local data (20\% data capacity) to achieve an average MAE of 3.52\%, outperforming the re-trained models even when trained on a full year of data (MAE of 3.55\%). Interestingly, \Cref{tab:capacity_results_aggregated} reveals that transferability strength varies significantly by meteorological cluster. For instance, Cluster 1 generalises rapidly to new sites, achieving an MAE of 3.7\% with only about 2.4 months of data. In contrast, Cluster 7 requires longer site-specific tuning, yielding an initial MAE of 5.8\% at 2.4 months, which improves to 3.4\% with 7.2 months. This suggests that while the source GPs successfully encode the broad dynamics of the respective weather patterns and take advantage of clusters that are easy to transfer, the meteorological phenomena associated with certain clusters are more location-dependent, necessitating a larger volume of local data for effective fine-tuning. Generally, adding data helps to reduce the variance in the fine-tuned location.
\begin{table}[ht]
\centering
\caption{Transfer learning performance by Cluster and Data Capacity, mean (std) across all target farms.}
\label{tab:capacity_results_aggregated}
\resizebox{\textwidth}{!}{
\begin{tabular}{c|cc|cc|cc|cc|cc}
\hline
 & \multicolumn{10}{c}{Data Capacity} \\
Clusters 
& \multicolumn{2}{c}{10\%} 
& \multicolumn{2}{c}{20\%} 
& \multicolumn{2}{c}{30\%} 
& \multicolumn{2}{c}{40\%} 
& \multicolumn{2}{c}{50\%} \\
\cline{2-3} \cline{4-5} \cline{6-7} \cline{8-9} \cline{10-11}
 & MAE & $R^2$ 
 & MAE & $R^2$ 
 & MAE & $R^2$ 
 & MAE & $R^2$ 
 & MAE & $R^2$ \\ 
\hline
\noalign{\vspace{0.5ex}}
0 & 4.4 (0.57) & 0.96 & 3.7 (0.41) & 0.97 & 3.5 (0.37) & 0.97 & 3.5 (0.33) & 0.97 & 3.5 (0.33) & 0.97 \\

1 & 3.7 (0.56) & 0.95 & 3.3 (0.35) & 0.96 & 3.2 (0.39) & 0.96 & 3.2 (0.32) & 0.96 & 3,1 (0.37) & 0.96 \\

2 & 3.9 (1.07) & 0.85 & 3.2 (0.79) & 0.89 & 3.0 (0.7) & 0.91 & 2.9 (0.76) & 0.91 & 2.9 (0.74) & 0.91 \\

3 & 4.5 (1.03) & 0.96 & 3.7 (0.73) & 0.97 & 3.3 (0.38) & 0.98 & 3.2 (0.43) & 0.98 & 3.2 (0.46) & 0.98 \\

4 & 5.1 (2.16) & 0.94 & 3.9 (0.76) & 0.97 & 3.5 (0.63) & 0.97 & 3.4 (0.51) & 0.97 & 3.4 (0.51) & 0.97 \\

5 & 4.5 (1.33) & 0.94 & 3.6 (0.77) & 0.95 & 3.2 (0.62) & 0.96 & 3.1 (0.56) & 0.97 & 3.1 (0.52) & 0.97 \\

6 & 4.8 (1.26) & 0.92 & 3.7 (0.63) & 0.96 & 3.4 (0.52) & 0.96 & 3,3 (0.4) & 0.96 & 3,3 (0.43) & 0.96 \\

7 & 5.8 (4.07) & 0.83 & 3.8 (1.21) & 0.93 & 3.4 (0.76) & 0.94 & 3.2 (0.56) & 0.95 & 3.1 (0.61) & 0.95 \\ 
\hline
\noalign{\vspace{1ex}}
TL framework Average 
& 4.34 (0.31) & 0.933 
& 3.52 (0.21) & 0.956
& 3.30 (0.20) & 0.960 
& 3.23 (0.18) & 0.963
& 3.22 (0.21) & 0.963 \\
\noalign{\vspace{1ex}}
\hline
\hline
\noalign{\vspace{1ex}}
Farm-level GP Average
& 3.84 (0.27) & 0.974 
& 3.87 (0.22) & 0.975 
& 3.78 (0.20) & 0.974 
& 3.63 (0.21) & 0.976 
& 3.56 (0.23) & 0.977 \\ 
\noalign{\vspace{1ex}}
\hline
\noalign{\vspace{1ex}}
Farm-level NN Average
& 4.53 (0.31) & 0.961 
& 3.89 (0.22) & 0.971 
& 3.71 (0.28) & 0.975 
& 3.41 (0.25) & 0.979 
& 3.51 (0.25) & 0.979 \\ 
\noalign{\vspace{1ex}}
\hline
\noalign{\vspace{1ex}}
Cluster-specific GPs Average 
& 5.26 (0.35) & 0.911 
& 4.13 (0.25) & 0.945 
& 3.74 (0.20) & 0.953 
& 3.59 (0.19) & 0.957 
& 3.55 (0.21) & 0.959 \\ 
\noalign{\vspace{1ex}}
\hline
\end{tabular}
}
\end{table}
Farm-level models allow a comparison with industrial baselines in \Cref{tab:farm_GP_det} and \Cref{tab:farm_mlp_det}. While single farm-level GPs outperform our framework with 10\% of data, our approach achieves lower MAE across all subsequent data capacities. Interestingly, the farm-level GP baseline shows minimal improvement between 10\% and 50\% data capacity, improving only from an MAE of 3.84\% to 3.56\%. This early plateau highlights the inherent ability of GP models to generalise quickly with limited data and explains the plateau in our framework's results between 40\% and 50\% capacity, when each cluster has more training samples. 

Probabilistic metrics in \Cref{tab:tab:farm_tl_proba} and \Cref{tab:farm_gp_proba} reveals our framework's uncertainty quantification. The CRPS mirrors the performance improvements in deterministic results and remains lower than the MAE across all data capacities. This highlights a strength of our approach: even when the GPs' predictive mean deviates from the actual observation, the probabilistic variance successfully expands to capture the true value. Furthermore, assessing the calibration of these bounds through the PICP$_{95}$ shows our framework achieves a near-optimal PICP$_{95}$ of 0.949 at 40\% data capacity. In contrast, the baseline GP yields a PICP$_{95}$ of 0.982 at 10\% capacity (which increases with capacity), indicating less informative intervals. Finally, when benchmarked against a NN, our approach maintains superior performance across all data capacities, underscoring the data efficiency and robust generalisation of our GP-based transfer method over more complex deep learning architectures.

\noindent \textbf{Farm-specific Analysis} Breaking down performance by target farms shows how transferability varies according to the meteorological conditions unique to each farm. As shown in \Cref{tab:capacity_results} for the Gemini wind farm, Clusters 0, 1, 2, 3, and 7 exhibit strong transferability. In these regimes, the pre-trained GP models align well with local dynamics, all yielding MAEs equal or below 3.4\% with approximately five months of fine-tuning data. Crucially, these clusters represent 66.7\% of the evaluated time periods (at 20\% data capacity), underscoring the operational benefit of the framework. Conversely, Clusters 4, 5, and 6 have lower transferability, suggesting that their associated weather patterns are specific to location, necessitating additional local data to refine the representations learned by the GPs.
\begin{table}[ht]
\centering
\caption{Transfer learning performance by Cluster and Data Capacity, mean(std), for the Gemini Wind Farm}
\label{tab:capacity_results}

\resizebox{\textwidth}{!}{
    \begin{tabular}{c|c|ccccc}
    \hline
     & & \multicolumn{5}{c}{Data Capacity} \\
    Clusters 
    & \multirow{2}{*}{\makecell{Test Samples \\ at 20\% capacity}}
    & 10\% 
    & 20\% 
    & 30\% 
    & 40\% 
    & 50\% \\
    \cline{3-7}
     & & MAE  
     & MAE
     & MAE  
     & MAE  
     & MAE \\
    \hline
    \noalign{\vspace{1ex}}
    0 & 1788 & 4.3 (0.26) & 3.3 (0.50) & 3,1 (0.60) & 3.3 (0.49) & 3.0 (0.40) \\
    1 & 2095 & 4.2 (0.51) & 3.4 (0.43) & 3.3 (0.44) & 3.3 (0.29) & 3.2 (0.40) \\
    2 & 1492 & 2.4 (1,16) & 2.2 (1,24) & 2.2 (1.28) & 2.1 (1.19) & 1.9 (1.00) \\
    3 & 1448 & 4.3 (1,09) & 3,4 (0.61) & 3.1 (0.20) & 3.2 (0.27) & 3.1 (0.26) \\
    4 & 1358 & 5.5 (2.66) & 4.4 (1,79) & 3.8 (1,08) & 3.6 (0.84) & 3.4 (0.52) \\
    5 & 1223 & 5.5 (1,81) & 3.9 (0.47) & 3.6 (0.44) & 3.4 (0.37) & 3.5 (0.30) \\
    6 & 1333 & 6.7 (2.39) & 4.2 (1.43) & 3.6 (0.92) & 3.4 (0.68) & 3,5 (0.83) \\
    7 & 922 & 4.4 (2.66) & 2.7 (1.08) & 2.7 (0.80) & 2.6 (0.84) & 2.5 (0.87) \\
    \hline
    \noalign{\vspace{1ex}}
    TL Framework Average 
    & 
    & 4.29 (0.14)
    & 3,31 (0.07)
    & 3.10 (0.03) 
    & 3.06 (0.03) 
    & 2.98 (0.03) \\
    \noalign{\vspace{1ex}}
    \hline
    \hline
    \noalign{\vspace{1ex}}
    Farm Level GP Average 
    & 
    & 3.5 (0.06)
    & 3,62 (0.21)
    & 3.46 (0.09)
    & 3.32 (0.05)
    & 3.27 (0.07) \\
    \noalign{\vspace{1ex}}
    \hline
    \noalign{\vspace{1ex}}
    Farm Level NN Average 
    & 
    & 4.29 (0.10)
    & 3.63 (0.08) 
    & 3.55 (0.01) 
    & 3.08 (0.32) 
    & 3.30 (0.11) \\
    \noalign{\vspace{1ex}}
    \hline    
    \noalign{\vspace{1ex}}
    Cluster-specific GPs Average
    & 
    & 5.06 (0.11)
    & 3,90 (0.02)
    & 3.53 (0.02) 
    & 3.35 (0.01) 
    & 3.31 (0.03) \\
    \noalign{\vspace{1ex}}
    \hline
    \end{tabular}
} 
\end{table}

\vspace{-0.1cm}

%% file: Chapters/Conclusion.tex
\label{sec:discussion}
The proposed framework demonstrates that weather-specific transfer learning reduces data requirements for accurate forecasting in European offshore wind farms. To scale this approach, future work will pursue two  main avenues. Firstly, scaling for longer lead times, which will require factorising the time series data, allowing for longer contexts that select key information for precise forecasting. This should pave the way for geographic transfer of clusters into more diverse climate zones by identifying the parts of the time series which will be crucial in forecasting future patterns, and thus overcoming the shortcoming of our approach for clusters that need longer fine-tuning. In a second avenue, using real-world data will bring multiple challenges. These include incorporating turbine specifications and wake effect modelling to account for intra-farm heterogeneity, as well as handling less granular data, since early-stage collection often contains missing data points and irregular observations. This research direction will counter the biased outcome of using synthetic data that does not necessarily correspond to operational environments. Additionally, implementing this methodology at production would require cross-validation of the weights given to each metric used to create the composite score, to avoid clustering errors. Collectively, these advancements pave the way toward a dynamic WRA tool capable of generating rapid, location- and configuration-specific forecasts without the need for lengthy initial data investment, thus accelerating future investments into offshore wind farms, and allowing the growth of cheaper, greener energy.

%% file: Chapters/AppendixA.tex
\textbf{Implementation details.} All code was run on a laptop (Lenovo YOGA S730, Intel Core i7-8565U, 16GB RAM). All results given are averaged for random seeds 42, 63, 84.

\noindent\textbf{Baseline Architecture.} For the neural network baseline detailed in \Cref{sec:results}, we employed a neural network with eight Multilayer Perceptron (MLP) hidden layers of dimensions 256, 256, 192, 128, 96, 64 and 32 that conclude with a single-neuron linear output. To ensure training stability and mitigate over-fitting, the first six hidden layers incorporate Layer Normalisation, a Gaussian Error Linear Unit (GELU) activation, and a 15\% dropout rate, while the penultimate 32-neuron layer utilises solely a GELU activation. The network was trained to minimise Mean Squared Error (MSE) using the Adam optimiser \citep{kingma2015adam} and has a learning rate of $10^{-3}$. Weight decay is set to $10^{-5}$, with a batch size of 256. Trained for 100 epochs.

\noindent\textbf{Pipeline implementation.} The full experimental pipeline was implemented in Python. Data handling used NumPy and pandas; train-test splits, scaling and pointwise error metrics used scikit-learn. The VAE and the NN-baseline were implemented in PyTorch; GP models used GPyTorch. Figures were produced with Matplotlib.

%% file: Chapters/AppendixB.tex
Model performance is assessed using a combination of deterministic point-forecast metrics and probabilistic metrics standard in the wind power forecasting literature and in the forecasting literature more generally.

\noindent\textbf{Mean Absolute Error} The Mean Absolute Error (MAE) quantifies the average magnitude of prediction errors without considering direction:
\begin{equation*}
    \text{MAE} = \frac{1}{n} \sum_{i=1}^{n} |y_i - \hat{y}_i|
\end{equation*}
where $y_i$ denotes the observed values output and $\hat{y}_i$ denotes the predicted values.

\noindent\textbf{Coefficient of Determination} The Coefficient of Determination ($R^2$) quantifies the proportion of variance in the observed data explained by the model:
\begin{equation*}
    R^2 = 1 - \frac{\sum_{i=1}^{n} (y_i - \hat{y}_i)^2}{\sum_{i=1}^{n} (y_i - \bar{y})^2}.
\end{equation*}
where $y_i$ denotes the observed values output, $\hat{y}_i$ denotes the predicted values and $\bar{y}$ denotes the mean of the observed values.

\noindent\textbf{Continuous Ranked Probability Score} 
The Continuous Ranked Probability Score (CRPS) is a strictly proper scoring rule that evaluates the calibration and sharpness of the entire predictive distribution. It is defined as the integrated squared difference between the cumulative distribution function (CDF) of the prediction and the empirical CDF of the observation:
\begin{equation*}
    \text{CRPS} = \frac{1}{n} \sum_{i=1}^{n} \int_{-\infty}^{\infty} (F_i(x) - H(x - y_i))^2 dx
\end{equation*}
where $F_i$ is the predictive CDF for the $i$-th observation, $y_i$ is the $i$-th actual observed value, and $H$ is the Heaviside step function ($H(z) = 1$ if $z \ge 0$, and $0$ otherwise). 

\noindent\textbf{Prediction-Interval Coverage Probability} 
The 95\% Prediction-Interval Coverage Probability (PICP$_{95}$) quantifies the empirical reliability of the model's uncertainty bounds. It calculates the proportion of true observations that successfully fall within the model's predicted 95\% prediction intervals:
\begin{equation*}
    \text{PICP}_{95} = \frac{1}{n} \sum_{i=1}^{n} \mathds{1}(L_i \le y_i \le U_i)
\end{equation*}
where $L_i$ and $U_i$ represent the respective lower and upper bounds of the 95\% predictive interval for the $i$-th observation.

%% file: Chapters/AppendixC.tex
To validate the robustness of the composite quality score $Q$, we performed a sensitivity analysis evaluating the impact of the time-period length, $p$, and number of clusters, $K$, on each constituent metric independently. This sensitivity analysis breaks down the contributions of each metric to the final composite score. Results are in \Cref{tab:all_metrics}.

\begin{table}[H]
    \centering
    \footnotesize
    \caption{Performance of cluster size and time period length configurations across each composite score metric.}
    \label{tab:all_metrics}

    \begin{subtable}[ht]{0.48\textwidth}
        \centering
        \caption{Only Silhouette Score}
        \begin{tabular}{c|ccccc}
            \toprule
            Time & \multicolumn{5}{c}{Clusters} \\
            Period & 8 & 10 & 12 & 14 & 16 \\
            \midrule
            6  & 0.168 & 0.143 & 0.132 & 0.137 & 0.129 \\
            12 & 0.190 & 0.175 & 0.160 & 0.143 & 0.136 \\
            24 & 0.168 & 0.153 & 0.145 & 0.138 & 0.133 \\
            36 & 0.147 & 0.133 & 0.124 & 0.126 & 0.128 \\
            48 & 0.162 & 0.152 & 0.147 & 0.149 & 0.144 \\
            \bottomrule
        \end{tabular}
    \end{subtable}
    \hfill
    \begin{subtable}[ht]{0.48\textwidth}
        \centering
        \caption{Only Meteorological Separability}
        \begin{tabular}{c|ccccc}
            \toprule
            Time & \multicolumn{5}{c}{Clusters} \\
            Period & 8 & 10 & 12 & 14 & 16 \\
            \midrule
            6  & 0.536 & 0.563 & 0.562 & 0.576 & 0.585 \\
            12 & 0.512 & 0.512 & 0.504 & 0.520 & 0.515 \\
            24 & 0.427 & 0.437 & 0.439 & 0.442 & 0.440 \\
            36 & 0.377 & 0.376 & 0.385 & 0.386 & 0.386 \\
            48 & 0.377 & 0.370 & 0.392 & 0.387 & 0.379 \\
            \bottomrule
        \end{tabular}
    \end{subtable}

    \vspace{0.5cm} 

    \begin{subtable}[ht]{0.48\textwidth}
        \centering
        \caption{Only Davies-Bouldin Index}
        \begin{tabular}{c|ccccc}
            \toprule
            Time & \multicolumn{5}{c}{Clusters} \\
            Period & 8 & 10 & 12 & 14 & 16 \\
            \midrule
            6  & 0.237 & 0.207 & 0.206 & 0.216 & 0.236 \\
            12 & 0.193 & 0.219 & 0.216 & 0.194 & 0.206 \\
            24 & 0.169 & 0.167 & 0.187 & 0.183 & 0.200 \\
            36 & 0.174 & 0.165 & 0.187 & 0.184 & 0.197 \\
            48 & 0.198 & 0.192 & 0.205 & 0.213 & 0.238 \\
            \bottomrule
        \end{tabular}
    \end{subtable}
    \hfill
    \begin{subtable}[ht]{0.48\textwidth}
        \centering
        \caption{Only Temporal Coherence}
        \begin{tabular}{c|ccccc}
            \toprule
            Time & \multicolumn{5}{c}{Clusters} \\
            Period & 8 & 10 & 12 & 14 & 16 \\
            \midrule
            6  & 0.782 & 0.813 & 0.819 & 0.826 & 0.834 \\
            12 & 0.713 & 0.720 & 0.737 & 0.757 & 0.766 \\
            24 & 0.592 & 0.625 & 0.650 & 0.667 & 0.685 \\
            36 & 0.554 & 0.564 & 0.581 & 0.595 & 0.603 \\
            48 & 0.492 & 0.511 & 0.545 & 0.561 & 0.574 \\
            \bottomrule
        \end{tabular}
    \end{subtable}

    \vspace{0.5cm}

    \begin{subtable}[ht]{0.48\textwidth}
        \centering
        \caption{Only Calinski-Harabasz}
        \begin{tabular}{c|ccccc}
            \toprule
            Time & \multicolumn{5}{c}{Clusters} \\
            Period & 8 & 10 & 12 & 14 & 16 \\
            \midrule
            6  & 0.933 & 0.919 & 0.909 & 0.902 & 0.894 \\
            12 & 0.842 & 0.829 & 0.820 & 0.811 & 0.803 \\
            24 & 0.766 & 0.752 & 0.740 & 0.731 & 0.724 \\
            36 & 0.708 & 0.695 & 0.683 & 0.676 & 0.670 \\
            48 & 0.703 & 0.690 & 0.678 & 0.669 & 0.662 \\
            \bottomrule
        \end{tabular}
    \end{subtable}
    \hfill
    \begin{subtable}[ht]{0.48\textwidth}
        \centering
        \caption{Distribution Consistency}
        \begin{tabular}{c|ccccc}
            \toprule
            Time & \multicolumn{5}{c}{Clusters} \\
            Period & 8 & 10 & 12 & 14 & 16 \\
            \midrule
            6  & 0.998 & 0.998 & 0.997 & 0.998 & 0.997 \\
            12 & 0.998 & 0.997 & 0.996 & 0.995 & 0.994 \\
            24 & 0.998 & 0.997 & 0.997 & 0.996 & 0.995 \\
            36 & 0.993 & 0.992 & 0.992 & 0.992 & 0.992 \\
            48 & 0.996 & 0.995 & 0.994 & 0.995 & 0.994 \\
            \bottomrule
        \end{tabular}
    \end{subtable}

\end{table}

%% file: Chapters/AppendixD.tex
Below is a breakdown of the probabilistic results of our framework (\Cref{tab:tab:farm_tl_proba}) as well as baseline results, stratified by cluster, data capacity and target wind farm where appropriate. Unless presented at the individual farm level (\Cref{tab:farm_GP_det}, \Cref{tab:farm_gp_proba} and \Cref{tab:farm_mlp_det}), the results are aggregated across all eight target farms (\Cref{tab:farm_GP_proba}). All reported metrics represent an average over three runs with MAE and CRPS displayed in \% of farm capacity values with mean(std).

\begin{table}[ht]
\centering
\caption{Probabilistic metrics of our proposed TL Framework}
\label{tab:tab:farm_tl_proba}

\resizebox{\textwidth}{!}{
\begin{tabular}{c|cc|cc|cc|cc|cc}
\hline
 & \multicolumn{10}{c}{Data Capacity} \\
Clusters 
& \multicolumn{2}{c}{10\%} 
& \multicolumn{2}{c}{20\%} 
& \multicolumn{2}{c}{30\%} 
& \multicolumn{2}{c}{40\%} 
& \multicolumn{2}{c}{50\%} \\
\cline{2-3} \cline{4-5} \cline{6-7} \cline{8-9} \cline{10-11}
 & CRPS & PICP$_{95}$ 
 & CRPS & PICP$_{95}$
 & CRPS & PICP$_{95}$ 
 & CRPS & PICP$_{95}$
 & CRPS & PICP$_{95}$ \\ 
\hline
\noalign{\vspace{0.5ex}}
0 & 3.4 (0.41) & 0.93 & 3.0 (0.27) & 0.94 & 2.9 (0.23) & 0.95 & 2.9 (0.21) & 0.95 & 2.9 (0.30) & 0.96 \\

1 & 3.0 (0.37) & 0.94 & 2.6 (0.28) & 0.93 & 2.6 (0.32) & 0.94 & 2.6 (0.27) & 0.96 & 2.6 (0.30) & 0.96 \\

2 & 3.2 (0.72) & 0.92 & 2.7 (0.57) & 0.93 & 2.6 (0.71) & 0.94 & 2.5 (0.54) & 0.95 & 2.5 (0.53) & 0.96 \\

3 & 3.5 (0.73) & 0.94 & 2.9 (0.51) & 0.94 & 2.7 (0.30) & 0.94 & 2.7 (0.31) & 0.95 & 2.7 (0.34) & 0.96 \\

4 & 3,9 (1.74) & 0.92 & 3.1 (0.50) & 0.94 & 2.8 (0.44) & 0.93 & 2.7 (0.35) & 0.94 & 2.8 (0.38) & 0.95 \\

5 & 3.4 (0.97) & 0.92 & 2.8 (0.56) & 0.92 & 2.6 (0.45) & 0.93 & 2.5 (0.42) & 0.93 & 2.5 (0.37) & 0.94 \\

6 & 3.7 (0.82) & 0.93 & 2.9 (0.43) & 0.93 & 2.7 (0.33) & 0.94 & 2.7 (0.26) & 0.95 & 2.7 (0.27) & 0.95 \\

7 & 4.7 (3.66) & 0.89 & 3.0 (0.86) & 0.93 & 2.8 (0.54) & 0.93 & 2.6 (0.43) & 0.93 & 2.6 (0.47) & 0.94 \\ 
\hline
\noalign{\vspace{1ex}}
Average 
& 3.41 (0.27) & 0.925 
& 2.82 (0.17) & 0.934
& 2.67 (0.15) & 0.936 
& 2.65 (0.15) & 0.949
& 2.55 (0.17) & 0.955 \\
\noalign{\vspace{1ex}}
\hline
\end{tabular}
}
\end{table}

\begin{table}[ht]
\centering
\caption{Deterministic cluster-specific GP baseline performance}
\label{tab:farm_GP_proba}

\resizebox{\textwidth}{!}{
\begin{tabular}{c|cc|cc|cc|cc|cc}
\hline
 & \multicolumn{10}{c}{Data Capacity} \\
Clusters 
& \multicolumn{2}{c}{10\%} 
& \multicolumn{2}{c}{20\%} 
& \multicolumn{2}{c}{30\%} 
& \multicolumn{2}{c}{40\%} 
& \multicolumn{2}{c}{50\%} \\
\cline{2-3} \cline{4-5} \cline{6-7} \cline{8-9} \cline{10-11}
 & MAE & $R^2$ 
 & MAE & $R^2$ 
 & MAE & $R^2$ 
 & MAE & $R^2$ 
 & MAE & $R^2$ \\ 
\hline
\noalign{\vspace{0.5ex}}
0 & 5.3 (0.83) & 0.94 & 4,4 (0.57) & 0.96 & 4.0 (0.41) & 0.97 & 3.9 (0.38) & 0.97 & 3.9 (0.36) & 0.97 \\

1 & 4.5 (0.65) & 0.93 & 3.8 (0.34) & 0.95 & 3.5 (0.35) & 0.96 & 3.5 (0.32) & 0.96 & 3,4 (0.40) & 0.96 \\

2 & 4.6 (1.18) & 0.81 & 3.7 (0.98) & 0.88 & 3.4 (0.83) & 0.89 & 3.2 (0.85) & 0.90 & 3.2 (0.84) & 0.91 \\

3 & 5.5 (1.43) & 0.95 & 4.4 (0.96) & 0.96 & 3.8 (0.47) & 0.97 & 3.6 (0.48) & 0.97 & 3.7 (0.68) & 0.97 \\

4 & 6.0 (2.23) & 0.93 & 4.6 (1.12) & 0.96 & 4.1 (0.80) & 0.97 & 3.9 (0.65) & 0.97 & 3.9 (0.61) & 0.97 \\

5 & 5.6 (1.78) & 0.90 & 4.2 (0.94) & 0.94 & 3.7 (0.67) & 0.95 & 3.5 (0.67) & 0.96 & 3.5 (0.60) & 0.96 \\

6 & 5.7 (1.28) & 0.90 & 4.3 (0.70) & 0.94 & 3.9 (0.63) & 0.95 & 3,6 (0.40) & 0.96 & 3,6 (0.43) & 0.96 \\

7 & 7.5 (4.29) & 0.78 & 4.6 (1.32) & 0.91 & 4.1 (0.88) & 0.92 & 3.7 (0.65) & 0.94 & 3.5 (0.65) & 0.94 \\ 
\hline
\noalign{\vspace{1ex}}
Average 
& 5.26 (0.35) & 0.911 
& 4.13 (0.25) & 0.945
& 3.74 (0.20) & 0.953 
& 3.59 (0.19) & 0.957
& 3.55 (0.21) & 0.959 \\
\noalign{\vspace{1ex}}
\hline
\end{tabular}
}
\end{table}

\begin{table}[ht]
\centering
\caption{Deterministic performance of farm-level GPs baseline}
\label{tab:farm_GP_det}

\resizebox{\textwidth}{!}{
\begin{tabular}{c|cc|cc|cc|cc|cc}
\hline
 & \multicolumn{10}{c}{Data Capacity} \\
Target Wind Farms
& \multicolumn{2}{c}{10\%} 
& \multicolumn{2}{c}{20\%} 
& \multicolumn{2}{c}{30\%} 
& \multicolumn{2}{c}{40\%} 
& \multicolumn{2}{c}{50\%} \\
\cline{2-3} \cline{4-5} \cline{6-7} \cline{8-9} \cline{10-11}
 & MAE & $R^2$ 
 & MAE & $R^2$ 
 & MAE & $R^2$ 
 & MAE & $R^2$ 
 & MAE & $R^2$ \\ 
\hline
\noalign{\vspace{0.5ex}}
Horns Rev & 3.7 (0.24) & 0.98 & 3.7(0.10) & 0.98 & 3.6(0.09) & 0.98 & 3.5(0.10) & 0.98 & 3.3(0.08) & 0.98 \\

Seagreen & 3.9 (0.12) & 0.97 & 4.1 (0.20) & 0.97 & 3.8 (0.07) & 0.98 & 3.6 (0.05) & 0.98 & 3,5 (0.13) & 0.98 \\

Gemini & 3.5 (0.06) & 0.97 & 3.6 (0.21) & 0.97 & 3.5 (0.09) & 0.97 & 3.3 (0.05) & 0.97 & 3.3 (0.07) & 0.97 \\

Hollandse Noord & 3.9 (0.17) & 0.97 & 3.7 (0.08) & 0.97 & 3.9 (0.08) & 0.97 & 3.8 (0.08) & 0.97 & 3.9 (0.06) & 0.97 \\

Dieppe & 4.3 (0.21) & 0.97 & 4.1 (0.06) & 0.98 & 3.9 (0.07) & 0.97 & 3.8 (0.07) & 0.98 & 3.8 (0.08) & 0.98 \\ 

Moray Firth & 3.6 (0.19) & 0.98 & 3.8 (0.05) & 0.98 & 3.6 (0.02) & 0.98 & 3.5 (0.02) & 0.98 & 3.4 (0.02) & 0.98 \\

Kriegers & 4.0 (0.14) & 0.97 & 4.1 (0.15) & 0.97 & 4.0 (0.17) & 0.97 & 3,9 (0.04) & 0.97 & 3,8 (0.04) & 0.97 \\

East Anglia One & 3.8 (0.14) & 0.98 & 3.8 (0.11) & 0.98 & 3.9 (0.06) & 0.97 & 3.6 (0.04) & 0.98 & 3.6 (0.12) & 0.98 \\ 
\hline
\noalign{\vspace{1ex}}
Average 
& 3.84 (0.27) & 0.974 
& 3.87 (0.22) & 0.975
& 3.78 (0.20) & 0.974
& 3.63 (0.21) & 0.976
& 3.56 (0.23) & 0.977 \\
\noalign{\vspace{1ex}}
\hline
\end{tabular}
}
\end{table}

\begin{table}[H]
\centering
\caption{Probabilistic performance of farm-level GPs baseline}
\label{tab:farm_gp_proba}

\resizebox{\textwidth}{!}{
\begin{tabular}{c|cc|cc|cc|cc|cc}
\hline
 & \multicolumn{10}{c}{Data Capacity} \\
Clusters 
& \multicolumn{2}{c}{10\%} 
& \multicolumn{2}{c}{20\%} 
& \multicolumn{2}{c}{30\%} 
& \multicolumn{2}{c}{40\%} 
& \multicolumn{2}{c}{50\%} \\
\cline{2-3} \cline{4-5} \cline{6-7} \cline{8-9} \cline{10-11}
 & CRPS & PICP$_{95}$ 
 & CRPS & PICP$_{95}$
 & CRPS & PICP$_{95}$ 
 & CRPS & PICP$_{95}$
 & CRPS & PICP$_{95}$ \\ 
\hline
\noalign{\vspace{0.5ex}}
Horns Rev & 3.2 (0.21) & 0.98 & 3.2 (0.11) & 0.99 & 3.3 (0.22) & 0.99 & 3.3 (0.16) & 0.99 & 3.2 (0.09) & 0.99 \\

Seagreen & 3.3 (0.09) & 0.98 & 3.2 (0.80) & 0.99 & 3.4 (0.09) & 0.99 & 3.4 (0.08) & 0.99 & 3.3 (0.10) & 0.99 \\

Gemini & 3.1 (0.14) & 0.98 & 3.2 (0.19) & 0.99 & 3.2 (0.14) & 0.99 & 3.1 (0.14) & 0.99 & 3.1 (0.11) & 0.99 \\

Hollandse Noord & 3.3 (0.10) & 0.98 & 3.2 (0.01) & 0.98 & 3.5 (0.05) & 0.99 & 3.5 (0.13) & 0.99 & 3.6 (0.14) & 0.99 \\

Dieppe & 3,7 (0.17) & 0.98 & 3.7 (0.26) & 0.99 & 3.7 (0.19) & 0.99 & 3.7 (0.20) & 0.99 & 3.6 (0.09) & 0.99 \\

Moray Firth & 3.1 (0.09) & 0.98 & 3,2 (0.02) & 0.98 & 3.3 (0.05) & 0.99 & 3.3 (0.04) & 0.99 & 3.3 (0.02) & 0.99 \\

Kriegers & 3.3 (0.14) & 0.98 & 3,6 (0.29) & 0.98 & 3.7 (0.06) & 0.99 & 3.7 (0.07) & 0.99 & 3.7 (0.04) & 0.99 \\

East Anglia One & 3.3 (0.05) & 0.99 & 3.3 (0.19) & 0.99 & 3.5 (0.06) & 0.99 & 3.3 (0.03) & 0.99 & 3.3 (0.11) & 0.99 \\ 
\hline
\noalign{\vspace{1ex}}
Average 
& 3.29 & 0.982 
& 3,33 & 0.986
& 3,45 & 0.992 
& 3.41 & 0.993
& 3.39 & 0.994 \\
\noalign{\vspace{1ex}}
\hline
\end{tabular}
}
\end{table}

\begin{table}[H]
\centering
\caption{Deterministic performance of farm-level MLPs baseline}
\label{tab:farm_mlp_det}

\resizebox{\textwidth}{!}{
\begin{tabular}{c|cc|cc|cc|cc|cc}
\hline
 & \multicolumn{10}{c}{Data Capacity} \\
Clusters 
& \multicolumn{2}{c}{10\%} 
& \multicolumn{2}{c}{20\%} 
& \multicolumn{2}{c}{30\%} 
& \multicolumn{2}{c}{40\%} 
& \multicolumn{2}{c}{50\%} \\
\cline{2-3} \cline{4-5} \cline{6-7} \cline{8-9} \cline{10-11}
 & MAE & $R^2$ 
 & MAE & $R^2$ 
 & MAE & $R^2$ 
 & MAE & $R^2$ 
 & MAE & $R^2$ \\ 
\hline
\noalign{\vspace{0.5ex}}
Horns Rev & 4.4 (0.19) & 0.96 & 3.9 (0.08) & 0.97 & 3.5 (0.07) & 0.98 & 3.3 (0.13) & 0.98 & 3.2 (0.15) & 0.98 \\

Seagreen & 4.5 (0.11) & 0.96 & 3.7 (0.03) & 0.97 & 3.6 (0.33) & 0.98 & 3.4 (0.24) & 0.98 & 3,5 (0.25) & 0.98 \\

Gemini & 4.3 (0.10) & 0.96 & 3.6 (0.08) & 0.97 & 3.5 (0.02) & 0.97 & 3.1 (0.32) & 0.98 & 3.3 (0.11) & 0.98 \\

Hollandse Noord & 4.5 (0.05) & 0.96 & 4.0 (0.14) & 0.97 & 3.9 (0.24) & 0.97 & 3.6 (0.26) & 0.98 & 3.7 (0.04) & 0.98 \\

Dieppe & 5.0 (0.11) & 0.96 & 4.2 (0.14) & 0.97 & 3.9 (0.25) & 0.97 & 3.6 (0.17) & 0.98 & 3.8 (0.06) & 0.98 \\

Moray Firth & 4.2 (0.12) & 0.97 & 3.7 (0.18) & 0.97 & 4.0 (0.20) & 0.98 & 3.5 (0.24) & 0.98 & 3.5 (0.04) & 0.98 \\

Kriegers & 5.0 (0.15) & 0.96 & 4.2 (0.22) & 0.97 & 3.9 (0.19) & 0.97 & 3,6 (0.05) & 0.98 & 3,5 (0.29) & 0.98 \\

East Anglia One & 4.3 (0.09) & 0.97 & 3.8 (0.16) & 0.97 & 3.4 (0.15) & 0.98 & 3.3 (0.03) & 0.98 & 3.5 (0.41) & 0.98 \\ 
\hline
\noalign{\vspace{1ex}}
Average 
& 4.53 (0.31) & 0.961 
& 3.89 (0.22) & 0.971
& 3.71 (0.28) & 0.975 
& 3.41 (0.25) & 0.979
& 3.51 (0.25) & 0.979 \\
\noalign{\vspace{1ex}}
\hline
\end{tabular}
}
\end{table}